\documentclass[runningheads]{llncs}
\usepackage[T1]{fontenc}
\usepackage{graphicx}
\usepackage{bm}
\usepackage{amsmath}
\usepackage{amssymb}
\usepackage{multirow}
\usepackage{subfigure}
\usepackage{makecell}
\usepackage{xspace}
\newcommand{\ourmethod}{CLEAR\xspace}

\begin{document}
\title{CLEAR: Cluster-based Prompt Learning \\ on Heterogeneous Graphs}
\titlerunning{Prompt Learning on Heterogeneous Graphs}
%
\author{Feiyang Wang$^1$, Zhongbao Zhang$^1$, Junda Ye$^1$, Li Sun$^2$, Jianzhong Qi$^2$}
\institute{$^1$Beijing University of Posts and Telecommunications, Beijing, China \\
$^2$North China Electric Power University, Beijing, China\\
\{fywang, zhongbaozb, jundaye\}@bupt.edu.cn, ccesunli@ncepu.edu.cn\\
jianzhong.qi@unimelb.edu.au
}
 \authorrunning{Wang, F., et al.}
%
\maketitle              

\begin{abstract}
Prompt learning has attracted increasing attention in the graph domain as a means to bridge the gap between pretext and downstream tasks. Existing studies on heterogeneous graph prompting typically use feature prompts to modify node features for specific downstream tasks, which do not concern the structure of heterogeneous graphs. Such a design also overlooks information from the meta-paths, which are core to learning the high-order semantics of the heterogeneous graphs. To address these issues, we propose \ourmethod, a \underline{C}luster-based prompt \underline{LEAR}ning model on heterogeneous graphs. We present cluster prompts that reformulate downstream tasks as heterogeneous graph reconstruction. In this way, we align the pretext and downstream tasks to share the same training objective. Additionally, our cluster prompts are also injected into the meta-paths such that the prompt learning process incorporates high-order semantic information entailed by the meta-paths. Extensive experiments on downstream tasks confirm the superiority of \ourmethod. It consistently outperforms state-of-the-art models, achieving up to 5\% improvement on the F1 metric for node classification.

\keywords{Heterogeneous graph \and Prompt learning \and Graph clustering.}
\end{abstract}
\section{Introduction}

Graphs are a core data structure in many applications, such as social networks \cite{GraphSAGE,GraphRec,sigir24sun}, citation networks \cite{Planetoid,GCN}, knowledge graphs \cite{TransE,RotatE,www22KGwang}, and bio-chemical graphs \cite{Protein,GTPN}. Graph neural networks (GNNs) have been applied to various graph learning problems, showing powerful results. In the real world, graphs are often formed by different types of nodes and edges, making ordinary GNNs inadequate for modeling such heterogeneous graphs. This leads to a series of studies on heterogeneous graph learning \cite{HetGNN,HAN}.

The ``pre-train, fine-tune'' paradigm is widely adopted to solve the problems of label scarcity and task dependence. Specifically, a graph model is pre-trained using self-supervised pretext tasks \cite{GPTGNN,GraphCL} and then fine-tuned with partial labels from downstream tasks. However, there is a gap between pretext and downstream tasks due to their different training objectives. Inspired by prompt learning in language models, researchers attempt to bridge the gap by employing the ``pre-train, prompt'' paradigm in the graph domain \cite{GPPT,All,www25sun}. These methods aim to use prompts to reformulate downstream tasks in line with the pretext task.

Although graph prompt learning has shown promising results, challenges remain when the technique is applied to heterogeneous graphs: 1) \textit{How to align the pretext task and downstream task through graph modification?} Existing heterogeneous graph prompting methods primarily design feature prompts to modify node features, allowing downstream tasks to focus on task-specific features. However, such prompts struggle to learn the structure information of heterogeneous graphs. 2) \textit{How to capture semantic information with prompts?} Meta-paths, as multi-hop connection patterns, reflect high-order semantics in heterogeneous graphs. Existing graph prompting methods typically leverage the direct neighbor connection to learn prompts, which overlooks the rich semantic information conveyed by meta-paths.

\begin{figure}[t]
    \centering
    \includegraphics[width=0.5\textwidth]{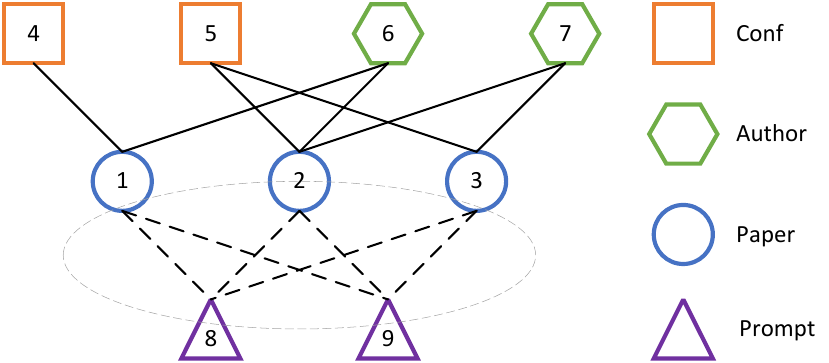}
    \caption{Cluster prompts in heterogeneous graphs. ``Paper'' is the target node and ``Prompt'' represents the cluster.}
    \label{figure::prompt}
    \vspace{-5mm}
\end{figure}

To address these limitations, we propose \ourmethod, a novel heterogeneous graph prompt learning model. Our intuition is that during graph training, nodes of the same label tend to learn embeddings that cluster together. We abstract each cluster as a virtual node that becomes our prompt. We insert the prompts into the original heterogeneous graph by connecting them to the target nodes. As heterogeneous graphs comprise multiple types of nodes and edges, we assign a new node type to the clusters and a new edge type to the connections. As shown in Figure \ref{figure::prompt}, ``Paper'' is the target node to be labeled. We regard each cluster as a prompt token and learn the adjacency matrix between ``Paper'' and ``Prompt''. Following the graph reconstruction above, downstream tasks such as node clustering and node classification can be reformulated as link prediction on the prompt-augmented heterogeneous graphs.

To capture both structural and semantic information, we design novel algorithms for both pre-training and prompt learning. During pre-training, we employ GNNs to learn structural and semantic features separately, and then use a contrastive framework to enhance the learning of both features. During prompt learning, we train prompts based on graph reconstruction and design a meta-path template to guide them in understanding high-order semantics. To bridge the gap between pretext and downstream tasks, we unify the optimization objective of pre-training and prompting. Specifically, we aggregate clustering features of target nodes from prompt tokens, which can be viewed as data augmentation w.r.t. structural or semantic features. Hence, the prompt tokens and the adjacency matrix can be learned using a contrastive loss aligned with pre-training.

The contributions of our work are summarized as follows:

\begin{itemize}
    \item We propose \ourmethod, a novel prompt learning model for heterogeneous graphs. To the best of our knowledge, this is the first attempt to reformulate downstream tasks as heterogeneous graph reconstruction with prompts.
    \item We introduce a contrastive framework to learn structural, semantic, and clustering information in heterogeneous graphs, which unifies the optimization objective of pre-training and prompting.
    \item We design a meta-path template to guide prompts in capturing semantics from meta-paths.
    \item Extensive experiments on downstream tasks demonstrate the superiority of our model against state-of-the-art heterogeneous graph learning models.
\end{itemize}

\section{Related Work}

\subsection{Graph Neural Networks}

Graph Neural Networks (GNNs) have emerged as powerful tools for learning representations of graph-structured data \cite{GCN,GAT,nips24sun}. Neighbor-based heterogeneous GNNs \cite{RGCN,HetGNN} generally capture the relational information from nodes of different types based on the connectivity of the graph. Path-based heterogeneous GNNs \cite{GTN,HGT,HAN,MAGNN}, on the other hand, focus on capturing high-order semantic information to learn node representations under the guidance of meta-paths.

\vspace{-2mm}

\subsection{Graph Pre-training}

Graph pre-training aims to mine graph information for downstream tasks in an unsupervised manner \cite{www25sun}. Recently, graph contrastive learning (GCL) \cite{GraphCL,GCC,GCA,aaai22selfMG,aaai23sun} has shown competitive performance in graph pre-training. The principle of GCL is to maximize the mutual information between positive sample pairs. For example, DMGI \cite{DMGI} maximizes the mutual information between local patches of a graph and the global representation of the entire graph. HeCo \cite{HeCo} introduces both network-schema and meta-path views for cross-view contrastive learning. SHGP \cite{SHGP} generates pseudo labels serving as self-supervised signals to guide node learning.

\vspace{-2mm}

\subsection{Prompt Learning}

As a powerful paradigm, prompt learning has attracted increasing attention with its flexibility and effectiveness. GPPT \cite{GPPT} introduces a token pair consisting of candidate label class and node entity, which reformulates the node classification task as edge prediction. Methods \cite{GPF,VNT-GPPE} inject feature prompts into node features, which can be optimized with few-shot labels for specific downstream tasks. To further relieve the difficulties of transferring prior knowledge to downstream domains, methods \cite{All,GraphPrompt} unify different downstream tasks as graph-level tasks with learnable prompts. HetGPT \cite{HetGPT} integrates a virtual class prompt and a heterogeneous feature prompt for heterogeneous graph learning. Further, HGPrompt \cite{HGPrompt} designs a unified graph prompt learning method for both heterogeneous and homogeneous graphs with dual templates and dual prompts.
\section{Preliminaries}

\begin{definition}{\textbf{Heterogeneous Graph.}}
    A heterogeneous graph is denoted as $\mathcal{G} = \{\mathcal{V},\mathcal{E}, \mathcal{T}, \mathcal{R}\}$, where $\mathcal{V}$ is the node set with a node type mapping function $\phi: \mathcal{V} \rightarrow \mathcal{T}$, and $\mathcal{E}$ is the edge set with an edge type mapping function $\varphi: \mathcal{E} \rightarrow \mathcal{R}$, respectively. Each node is associated with a node type $\phi(v) \in \mathcal{T}$ and each edge with a relation $\varphi(e) = \varphi(v_i,v_j) \in \mathcal{R}$. A heterogeneous graph can be represented by a set of adjacency matrices $\{\bm A_k\}_{k=i}^K$, where $K=|\mathcal{R}|$ is the number of edge types. $\bm A_k \in \mathbb{R}^{N \times N}$ denotes the adjacency matrix of the $k$-th type edge, where $N$ is the number of nodes in $\mathcal{V}$.
\end{definition}

\begin{definition}{\textbf{Meta-path.}}
    A meta-path $\mathcal{P}$ describes a pattern of connections in the heterogeneous graph, denoted as $T_1 \stackrel{R_1}{\longrightarrow} T_2 \stackrel{R_2}{\longrightarrow} \cdots \stackrel{R_l}{\longrightarrow} T_{l+1}$ (simplified to $T_1 T_2 \cdots T_{l+1}$), where $T_1,T_2,\cdots,T_{l+1} \in \mathcal{T}$ are node types, and $R_1,R_2,\cdots,R_l \in \mathcal{R}$ are relations.
\end{definition}

\section{\ourmethod}

\begin{figure}
	\centering
	\includegraphics[width=1\textwidth]{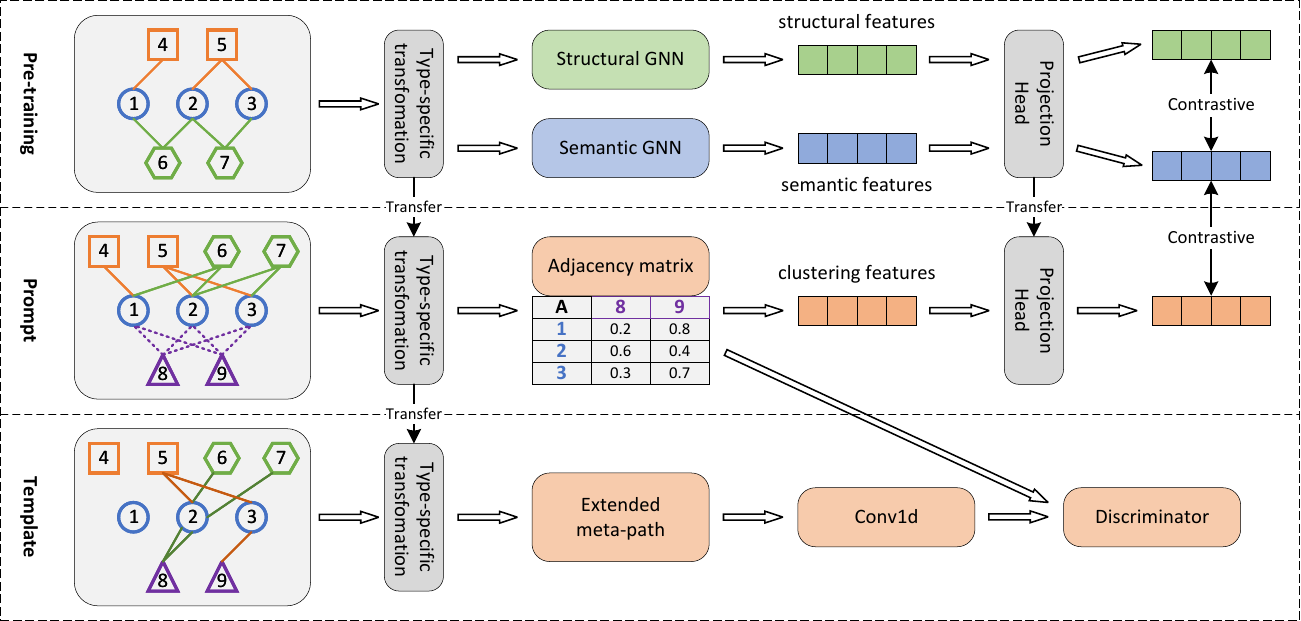}
	\caption{Overall framework of \ourmethod. The pre-training module learns structural and semantic features of heterogeneous graphs within a contrastive framework. The prompt module introduces cluster prompts to learn clustering features, also trained within the same contrastive framework. The template module integrates prompts into meta-paths to enhance prompt learning with high-order semantics. Parameters are transferred from the pre-training module to the prompt module and template module.}
	\label{figure::framework}
\end{figure}

We proposed a model named \ourmethod to bridge the gap between pre-training and downstream tasks on heterogeneous graphs. Our model has three main modules, including pre-training, cluster prompt, and meta-path template. The overall framework is shown in Figure \ref{figure::framework}.

\subsection{Pre-training on Heterogeneous Graph}\label{sec::pretrain}

Heterogeneous graphs contain not only structural information of node connections but also semantic information of meta-paths. To capture both structural and semantic features, we propose a self-supervised heterogeneous graph learning framework that separately acquires these features and fuses them using contrastive learning.

Due to the heterogeneity of the graph, the features of different types of nodes are distributed in different spaces. For easier message-passing between different types of nodes, we design type-specific transformation matrix $\bm M_{\phi}$ to project features into the common space:
\begin{equation}\label{equ::transformation}
    \bm h_i = \bm M_\phi \bm x_i,
\end{equation}
where $\bm x_i$ and $\bm h_i$ are the original and projected features of node $v_i$, respectively.

Assume that the target nodes of type $T_\tau$ are connected to neighbors of types $\{T_1,T_2,\cdots,T_K\} \in \mathcal{T}$. To aggregate node features from different types of neighbors, we present graph convolutional networks with the following type-wise propagation rule:
\begin{equation}
    \bm H_\tau^k = \sigma \left(\tilde{\bm A}_k \bm H_k \bm W_k\right),
\end{equation}
where $\tilde{\bm A}_k \in \mathbb{R}^{N_\tau \times N_k}$ is the normalized adjacency matrix from the given type $T_k$ to the target type $T_\tau$, $N_\tau$ and $N_k$ are the number of nodes of the corresponding type, $\bm H_k$ represents the set of node features of type $T_k$, and $\bm W_k$ is a type-specific trainable weight matrix. After aggregating features from different types, we employ the attention mechanism to measure the contribution of each type of nodes to the target nodes:
\begin{equation}\label{equ::structure}
    \bm H_\tau^{\mathcal{G}} = \sum_{k=1}^{K} \bm\alpha_k \bm H_\tau^k,
\end{equation}
where $\bm\alpha_k$ is the attention weight of type $T_k$:
\begin{equation}
    \alpha_k = \frac{\exp(\text{LeakyReLU}(\bm a^\top [\bm h_i||\bm h_i^k]))}{\sum_{l \in \mathcal{T}} \exp(\text{LeakyReLU}(\bm a^\top [\bm h_i||\bm h_i^l]))}.
\end{equation}
Here, $\bm a \in \mathbb{R}^{2d \times 1}$ is a learnable attention vector and $||$ denotes the concatenate operation.

In addition to learning structural information, we also utilize meta-paths to capture high-order semantic information of heterogeneous graphs. Given a set of meta-paths $\{\mathcal{P}_1,\mathcal{P}_2,\cdots,\mathcal{P}_M\}$, we can get a meta-path based adjacency matrix $\bm A_\tau \in \mathbb{R}^{N_\tau \times N_\tau}$ for the target node type. For neighbors that exist in multiple meta-paths (i.e., $v_i,v_j \in \mathcal{P}_1,\mathcal{P}_2$), we set the element of adjacency matrix to the corresponding count (i.e., $\bm A_{ij}=2$). Then we utilize a GCN layer to obtain the semantic feature of nodes:
\begin{equation}\label{equ::semantics}
    \bm H_\tau^{\mathcal{P}} = \sigma \left(\tilde{\bm A}_\tau \bm H_\tau \bm W_\tau\right),
\end{equation}
where $\tilde{\bm A}_\tau$ is the normalized adjacency matrix, $\bm W_\tau$ is a trainable weight matrix, and $\sigma$ is the sigmoid activation function.

After obtaining node representations of two views $\bm H_\tau^\mathcal{G}$ and $\bm H_\tau^\mathcal{P}$, we use a shared projection head $g(\cdot)$ to map these representations into the contrastive space:
\begin{equation}\label{equ::project}
	\bm Z = g(\bm H), \; \bm H \in \{\bm H_\tau^\mathcal{G},\bm H_\tau^\mathcal{P}\},
\end{equation}
where $g(\cdot)$ can be fully connected layers with activation functions. A contrastive loss $\mathcal{L}_{pre}$ is applied to pre-train the parameters by maximizing the agreement between $\bm Z^{\mathcal{G}}$ and $\bm Z^{\mathcal{P}}$.
\begin{equation}\label{equ::pretrain}
    \mathcal{L}_{pre} = -\sum_{i=1} \log \frac{\exp(\text{sim}(\bm z_i^{\mathcal{G}},\bm z_i^{\mathcal{P}}))}{\sum_{j=1}\exp(\text{sim}(\bm z_i^{\mathcal{G}},\bm z_j^{\mathcal{P}}))},
\end{equation}
where $z_i^{\mathcal{G}} \in \bm Z^{\mathcal{G}}$ and $\bm z_i^{\mathcal{P}} \in \bm Z^{\mathcal{P}}$, and $\text{sim}(\cdot,\cdot)$ is a similarity function (e.g., cosine similarity or Minkowski distance).

\subsection{Cluster Prompt}\label{sec::prompt}

The prompt aims to guide the pre-trained model to understand and predict the downstream task. In natural language processing, prompts are mostly in the form of cloze and prefix. Such prompts work well because words can naturally constitute phrases or sentences with suggestive semantics. For graphs, there are fixed relations and structures between nodes, the semantics of which cannot be expressed easily. Instead, existing homogeneous graph prompting methods \cite{GPPT,All} introduce learnable tokens to build prompts automatically. Nevertheless, this is infeasible for heterogeneous graphs due to their multiple node/edge types and complex structure.

To bridge the gap between pre-training and downstream tasks, we consider learning the node class based on structural clustering. As shown in Figure \ref{figure::prompt}, each cluster is treated as a learnable prompt token and fully connected to the target node with a trainable adjacency matrix. Thus, the node classification task can be viewed as a link prediction task between the target node and the prompt token.

In line with the structure of the heterogeneous graph, the prompt tokens are assigned a new node type $T_\rho$, and the type set is updated as $\mathcal{T} \leftarrow \mathcal{T} \cup \{T_\rho\}$. The clustering feature of target nodes is calculated as: 
\begin{equation}\label{equ::cluster}
	\bm H_\tau^\rho = \tilde{\bm A}_\rho \bm H_\rho,
\end{equation}
where $\bm H_\rho$ is the representation of prompt tokens and $\bm A_\rho \in \mathbb{R}^{N^\tau*N^\rho}$ is the adjacency matrix between prompt tokens and target nodes.

Consistent with the pre-training stage, we map the clustering feature into the contrastive space via the trained projection head $g(\cdot)$ and learn representations for the prompt tokens using the contrastive loss:
\begin{equation}
	\bm Z^{\rho} = g(\bm H_\tau^\rho),
\end{equation}
\begin{equation}\label{equ::prompt}
	\mathcal{L}_1 = -\sum_{i=1} \log \frac{\exp(\text{sim}(\bm z_i,\bm z_i^\rho))}{\sum_{j=1}\exp(\text{sim}(\bm z_i,\bm z_j^\rho))},
\end{equation}
where $\bm Z \in \{\bm Z^\mathcal{G},\bm Z^\mathcal{P}\}$ is the pre-trained projection of target nodes.

Moreover, in order to guarantee that the clusters are separated from each other, we utilize orthogonal constraint \cite{GPPT} to initialize and continuously learn prompt tokens:
\begin{equation}\label{equ::orthogonal}
    \mathcal{L}_0 = ||\bm H_\rho \bm H_\rho^\top - \bm I||_2^2,
\end{equation}
where $\bm I$ is the identity matrix.

\subsection{Meta-path Template}\label{sec::template}

In addition to direct neighbor relationships, meta-paths describe high-order semantic relationships. To further enhance node clustering with semantic information, we design a meta-path prompt template with an adjacency-guided discriminator. Specifically, given a meta-path ``XXTXX'' with the target node ``T'', we duplicate ``T'' and insert the prompt token ``P'' to construct the prompt template ``XX\underline{TPT}XX'', which also conforms to the original meta-path pattern. For instance, in a citation network, the meta-path ``ATA'' (i.e., author-paper-author, where paper is the target node type) represents the semantics of co-authorship. In order for the prompt token to understand this semantics, we generate an extended meta-path ``A\underline{TPT}A'' based on the prompt template.

After constructing the meta-path template, we introduce a discriminator to measure the correctness of generated meta-paths. Considering the specific connection pattern in meta-paths, we employ a one-dimensional convolutional neural network (Conv1d) to capture path semantics since it has a strong capability to process sequential data. A multilayer perceptron (MLP) with a sigmoid function serves as a probability distribution function to calculate the validity probability of the generated meta-paths:
\begin{equation}\label{equ::discriminator}
     \mathcal{D}(\mathcal{P}) = \sigma(\text{MLP}(\text{Conv1d}(\mathcal{P}))).
\end{equation}

For an original meta-path containing the target node, there are $N^\rho$ generated template paths to be discriminated, where $N^\rho$ is the number of prompt tokens. Here, we introduce an adjacency-based sampling method that measures the impact coefficient of different template paths. Specifically, the prompt token with the highest connection probability to the target node is regarded as the ground truth and the others are the negative samples. The loss function is given as:
\begin{equation}\label{equ::template}
    \mathcal{L}_2 = - \left( \sum_{\mathcal{P}_+} q_{\mathcal{P}_+} \log\mathcal{D}(\mathcal{P}_+) + \sum_{\mathcal{P}_-} q_{\mathcal{P}_-} \log(1-\mathcal{D}(\mathcal{P}_-)) \right),
\end{equation}
where $\mathcal{P}^+$ and $\mathcal{P}^-$ are positive and negative template paths, and the coefficient $q_{\mathcal{P}}$ is the connection probability between the prompt token and the target node in terms of the normalized adjacency matrix $\tilde{\bm A}_p$. By considering the impact of different paths, we train the prompt token in a self-adversarial manner, which improves the robustness and effectiveness of the model.


\subsection{Overall Learning Process}

Our prompting method can work under zero-shot or few-shot settings. In the pre-training stage, we learn node representations and model parameters using a contrastive loss. In the prompting stage, we fix the pre-trained parameters and optimize the prompt tokens $\bm H_\rho$ and the adjacency matrix $\bm A_\rho$. The loss function of zero-shot learning is given as:
\begin{equation}\label{equ::overall}
	\mathcal{L}_{prompt} = \alpha \mathcal{L}_1 + (1-\alpha) \mathcal{L}_2 + \beta \mathcal{L}_0,
\end{equation}
where $\alpha,\beta$ are hyper-parameters.

With regard to few-shot learning, the downstream tasks are given limited labels. We minimize the distance between the target node $v_i$ and its ground truth label $y_i$ while maximizing the distance between negative pairs:
\begin{equation}
	\mathcal{L}_{label} = \sum_i \left( d(\bm h_i^\tau,\bm h_{y_i}^\rho) - \mathbb{E}_{\{\phi(y) = T_\rho, y \neq y_i\}} d(\bm h_i^\tau,\bm h_{y}^\rho) \right),
\end{equation}
where $\bm h_i^\tau \in \bm H_\tau$ is the representation of the target node, $\bm h_{y_i}^\rho \in \bm H_\rho$ is the representation of the corresponding prompt token, and $d(\cdot,\cdot)$ is the Euclidean distance. Finally, the loss function of few-shot learning is given as: $\mathcal{L}_{few} = \mathcal{L}_{prompt} + \gamma \mathcal{L}_{label}$, where $\gamma$ is a hyper-parameter.


\section{Experiments}

\begin{table}[b]
    \caption{Statistics of the datasets}
    \centering
    \resizebox{0.8\textwidth}{!}{
    \setlength{\tabcolsep}{4mm}
    \begin{tabular}{c|c|c|c|c}
        \hline
        Dataset&\# Nodes&\# Edges&Meta-path&\# Classes\\ \hline
        ACM&\makecell{Paper: 4019\\Author: 7167\\Subject: 60}&\makecell{P-A: 13407\\P-S: 4019}&\makecell{PAP\\PSP}&\makecell{Paper\\3}\\ \hline
        DBLP&\makecell{Author: 4057\\Paper: 14328\\Term: 7723\\Conference: 20}&\makecell{P-A: 19645\\P-T: 85810\\P-C: 14328}&\makecell{APA\\APCPA\\APTPA}&\makecell{Author\\4}\\ \hline
        IMDB&\makecell{Movie: 4278\\Director: 2081\\Actor: 5257}&\makecell{M-D: 4278\\M-A: 12828}&\makecell{MAM\\MDM}&\makecell{Movie\\3}\\ \hline
    \end{tabular}}
    \label{tab::statistics}
\end{table}

\subsection{Experimental Setup}

\noindent\textbf{Datasets.} We evaluate \ourmethod on three benchmark datasets: ACM, DBLP, and IMDB \cite{HAN,MAGNN}. Detailed statistics of these datasets are given in Table \ref{tab::statistics}.

\noindent\textbf{Baselines.} We evaluate \ourmethod against the state-of-the-art models that are grouped into three main categories as follows. (1) Supervised learning: HetGNN \cite{HetGNN}, HAN \cite{HAN}, HGT \cite{HGT}, and MAGNN \cite{MAGNN}. (2) Pre-training: DMGI \cite{DMGI}, HeCo \cite{HeCo}, and SHGP \cite{SHGP}. (3) Prompt learning: HGPrompt \cite{HGPrompt} and HetGPT \cite{HetGPT}.

\begin{table}[t]
    \centering
    \caption{Evaluation results on node clustering}
    \resizebox{0.75\textwidth}{!}{
    \setlength{\tabcolsep}{3mm}
    \begin{tabular}{c|cc|cc|cc}
        \hline
        Dataset&\multicolumn{2}{c|}{ACM}&\multicolumn{2}{c|}{DBLP}&\multicolumn{2}{c}{IMDB}\\
        \hline
        Metric&NMI&ARI&NMI&ARI&NMI&ARI\\
        \hline
        DMGI&45.64&40.58&63.25&65.10&3.41&2.72\\
        HeCo&49.20&46.91&67.49&71.74&4.18&3.63\\
        SHGP&51.14&47.83&70.02&73.56&4.31&3.87\\
        HGPrompt&47.36&43.11&67.62&70.13&3.05&2.49\\
        HetGPT&52.58&49.37&70.42&72.61&4.13&3.65\\
        \textbf{\ourmethod}&\textbf{55.16}&\textbf{52.62}&\textbf{74.49}&\textbf{76.56}&\textbf{4.82}&\textbf{4.63}\\
        \hline
    \end{tabular}}
    \label{tab::clustering}
\end{table}

\begin{table}[t]
    \centering
    \caption{Evaluation results on 1-shot node classification}
    \resizebox{1\textwidth}{!}{
    \begin{tabular}{c|c|cc|cc|cc}
        \hline
        \multirow{2}{*}{Category}&Dataset&\multicolumn{2}{c|}{ACM}&\multicolumn{2}{c|}{DBLP}&\multicolumn{2}{c}{IMDB}\\
        &Metric&Macro-F1&Micro-F1&Macro-F1&Micro-F1&Macro-F1&Micro-F1\\
        \hline
        \multirow{4}{*}{Supervised Learning}&HetGNN&32.73$\pm$3.1&43.10$\pm$2.9&44.05$\pm$2.5&48.66$\pm$2.4&20.16$\pm$1.6&27.72$\pm$1.4\\
        &HAN&37.92$\pm$3.2&48.54$\pm$3.1&49.25$\pm$2.8&51.33$\pm$2.5&25.06$\pm$1.7&36.17$\pm$1.4\\
        &MAGNN&40.67$\pm$1.9&51.10$\pm$2.3&54.21$\pm$1.3&52.75$\pm$2.0&34.30$\pm$1.2&37.68$\pm$1.3\\
        &HGT&47.49$\pm$3.4&55.30$\pm$3.3&60.81$\pm$2.9&63.41$\pm$2.7&28.39$\pm$2.2&35.05$\pm$1.9\\
        \hline
        \multirow{3}{*}{Pre-training}&DMGI&44.10$\pm$1.5&50.35$\pm$1.4&74.89$\pm$2.2&75.46$\pm$2.0&30.02$\pm$1.8&33.21$\pm$1.7\\
        &HeCo&54.76$\pm$1.8&57.50$\pm$1.7&82.32$\pm$1.3&82.41$\pm$1.3&29.13$\pm$1.9&34.57$\pm$1.6\\
        &SHGP&59.01$\pm$1.3&63.29$\pm$1.3&84.74$\pm$1.2&85.32$\pm$1.1&36.83$\pm$1.2&38.25$\pm$1.3\\
        \hline
        \multirow{3}{*}{Prompting}&HGPrompt&60.35$\pm$1.9&62.92$\pm$2.2&82.54$\pm$1.8&84.76$\pm$2.0&34.31$\pm$1.6&37.25$\pm$1.7\\
        &HetGPT&63.97$\pm$1.6&65.24$\pm$1.8&85.14$\pm$1.2&85.60$\pm$1.2&39.19$\pm$1.9&40.06$\pm$1.8\\
        &\textbf{\ourmethod}&\textbf{69.19}$\pm$1.4&\textbf{71.43}$\pm$1.6&\textbf{88.86}$\pm$0.8&\textbf{89.07}$\pm$1.1&\textbf{44.50}$\pm$0.9&\textbf{45.48}$\pm$1.0\\
        \hline
    \end{tabular}}
    \label{tab::classification}
\end{table}

\noindent\textbf{Settings.} We consider two downstream tasks, i.e., node clustering and node classification. The node clustering task is learned under a zero-shot setting. We use the clustering feature of target nodes as the node embedding, and apply $K$-means algorithm to cluster the target nodes. The clustering quality is measured by normalized mutual information (NMI) and adjusted rand index (ARI). The node classification task is configured in $k$-shot learning, where $k$ ranges in $\{1,5,10,20,50\}$. We adopt Macro-F1 and Micro-F1 as the evaluation metrics.

\subsection{Main Results}

\noindent\textbf{Node Clustering}.
The result of zero-shot node clustering is reported in Table \ref{tab::clustering}. We omit the supervised models for comparison due to their data leakage during training. We can see that \ourmethod achieves the best results consistently over all datasets. Compared with pre-training models, \ourmethod gains an improvement of approximately 5\% on both metrics for the ACM and DBLP datasets, which demonstrates the effectiveness of our prompting model.

\noindent\textbf{Node Classification}.
The result of one-shot node classification is reported in Table \ref{tab::classification}. Compared with state-of-the-art prompting models, \ourmethod achieves an improvement of 5.6\%, 3.7\%, and 5.3\% in Macro-F1 and 6.2\%, 3.5\%, and 5.4\% in Micro-F1, respectively, demonstrating the superiority of our model. Under the one-shot learning setting, prompting models generally perform better than pre-training models and significantly better than supervised models. This indicates that when training labels are extremely scarce, pre-training and prompting models can better exploit inherent graph information. Moreover, the prompting models bridge the gap between pre-training and downstream tasks, leading to substantial performance gains.


\begin{figure}[t]
    \centering
    \includegraphics[width=0.3\textwidth]{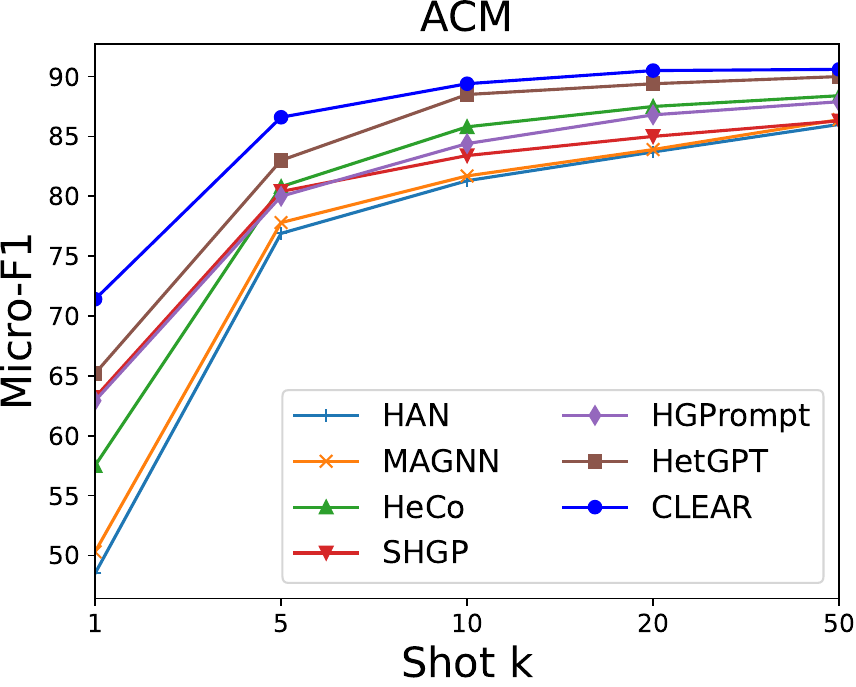}
    \includegraphics[width=0.3\textwidth]{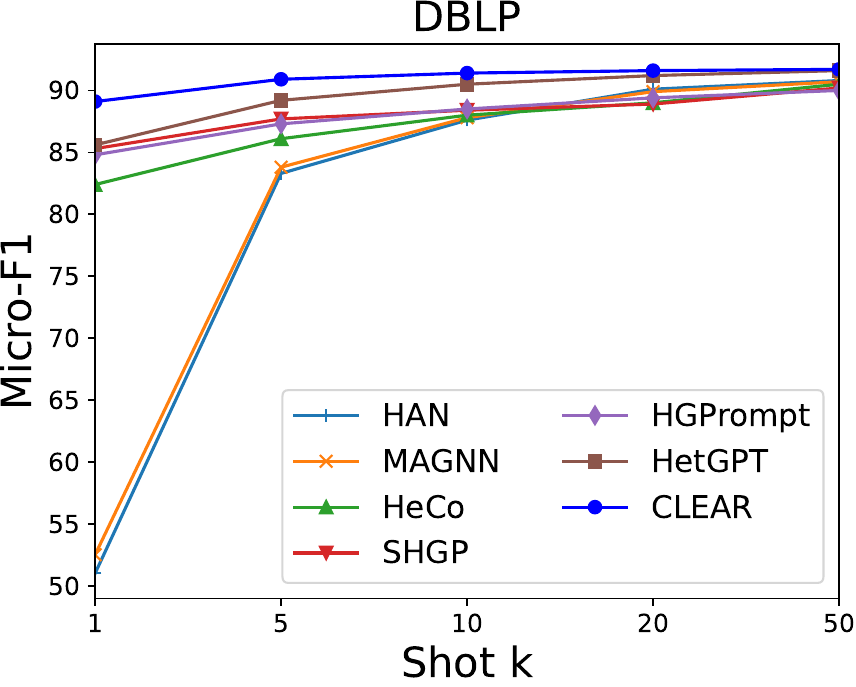}
    \includegraphics[width=0.3\textwidth]{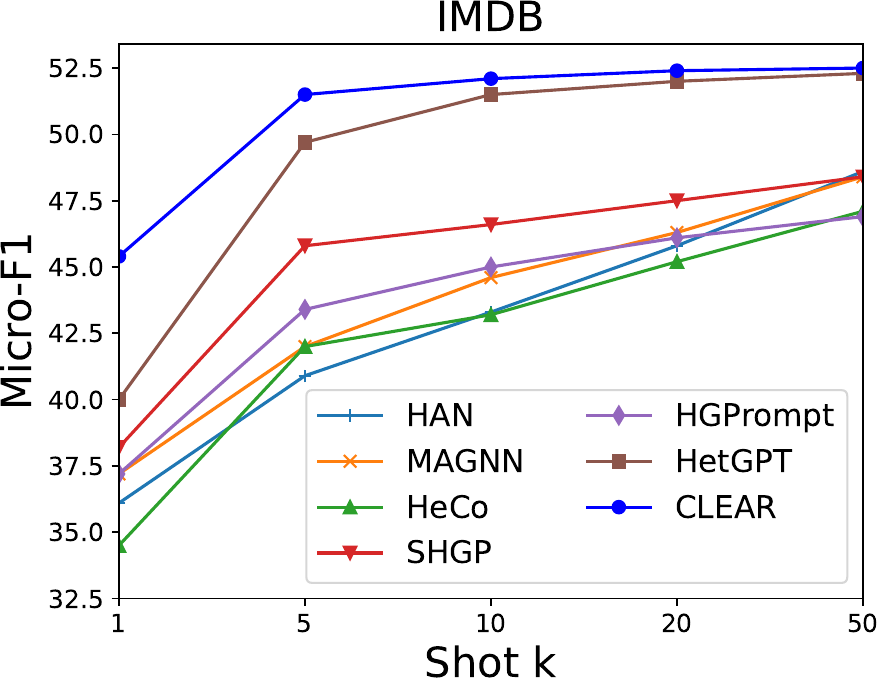}
    \caption{Model performance w.r.t. number of shots on node classification task}
    \label{fig::classification}
\end{figure}

\subsection{Shot Performance}
We vary the number of shots and plot the result in Figure \ref{fig::classification}. \ourmethod consistently outperforms other models. With fewer shots, the performance gap is greater, demonstrating the robustness of \ourmethod for few-shot learning. Compared with the ``pre-train, fine-tune'' paradigm, our ``pre-train, prompt'' model successfully bridges the gap between node classification and the pretext task.

\begin{figure}[t]
    \centering
    \subfigure[raw features]{\includegraphics[width=0.3\textwidth]{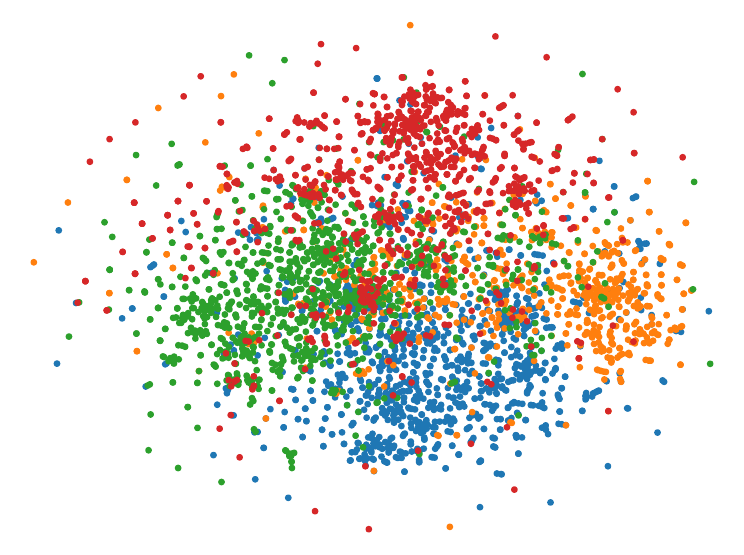}}
    \subfigure[pre-trained features]{\includegraphics[width=0.3\textwidth]{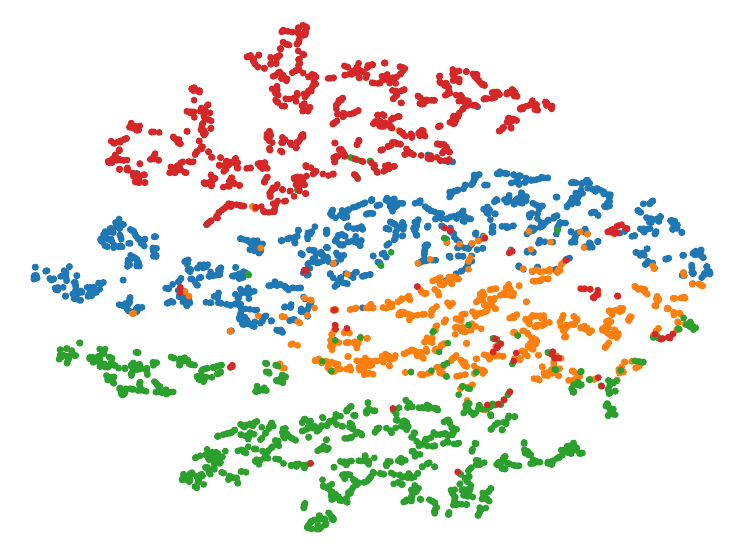}}
    \subfigure[clustering features]{\includegraphics[width=0.3\textwidth]{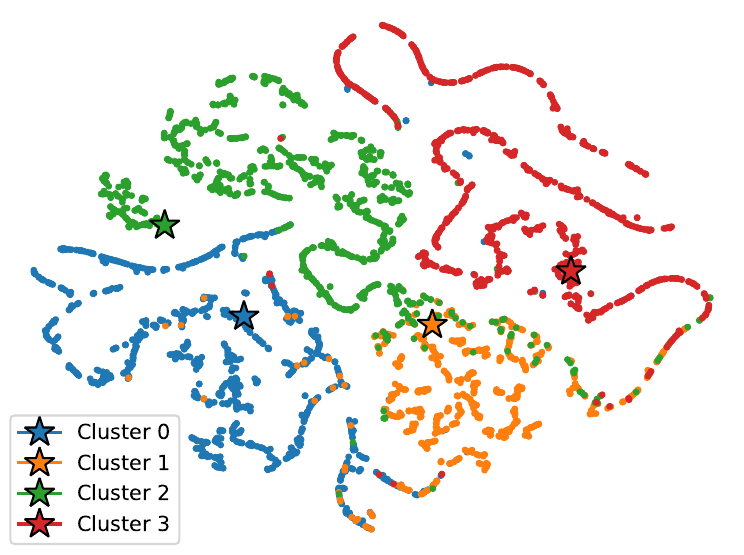}}
    \vspace{-3mm}
    \caption{Visualization of node embeddings under the zero-shot setting on DBLP}
    \label{fig::visualization}
\end{figure}

\subsection{Visualization}

To better understand the effectiveness of our prompting framework under the zero-shot setting, we plot nodes and prompt tokens in 2-dimensional Euclidean space using the t-SNE algorithm in Figure 4. Nodes are colored by their ground-truth labels, and prompt tokens are denoted by stars. We observed that after pre-training, nodes with different labels have been preliminarily separated, but there is still a gap in application to downstream tasks. In contrast to pre-trained features, the clustering features are orthogonally distributed and the prompt tokens are located within the clusters of nodes, which simplifies the application of downstream tasks.
\section{Conclusion}

We proposed a cluster-based heterogeneous graph prompting method named \ourmethod. It mainly comprises three modules: pre-training, cluster prompt, and meta-path template. The pre-training module applies contrastive learning to pre-train the graph, capturing both structural and semantic information in heterogeneous graphs. The cluster prompt module introduces cluster prompts to bridge the gap between pre-training and downstream tasks. The meta-path template module designs a meta-path template to strengthen prompt learning with high-order semantics. Extensive experiments on the node clustering and node classification tasks show that our method bridges the gap between pretext and downstream tasks and achieves superior results.

\bibliographystyle{splncs04}
\bibliography{ref}

\end{document}